\documentclass[letterpaper, 10 pt, conference]{ieeeconf}  

\IEEEoverridecommandlockouts                              

\usepackage{graphics} 
\usepackage{epsfig} 
\usepackage{mathptmx} 
\usepackage{times} 
\usepackage{amsmath} 
\usepackage{amssymb}  
\usepackage{booktabs} 
\usepackage[table,dvipsnames]{xcolor} 

\title{\LARGE \bf
GaussianSSC: Triplane-Guided Directional Gaussian Fields for 3D Semantic Completion
}

\author{Ruiqi Xian, Jing Liang, He Yin, Xuewei Qi, Dinesh Manocha
}

\begin{document}

\maketitle
\thispagestyle{empty}
\pagestyle{empty}


\begin{abstract}
We present \emph{GaussianSSC}, a two-stage, grid-native and triplane-guided approach to semantic scene completion (SSC) that injects the benefits of Gaussians without replacing the voxel grid or maintaining a separate Gaussian set. We introduce \emph{Gaussian Anchoring}, a sub-pixel, Gaussian-weighted image aggregation over fused FPN features that tightens voxel--image alignment and improves monocular occupancy estimation. We further convert point-like voxel features into a learned per-voxel Gaussian field and refine triplane features via a triplane-aligned \emph{Gaussian--Triplane Refinement} module that combines \emph{local gathering} (target-centric) and \emph{global aggregation} (source-centric). This directional, anisotropic support captures surface tangency, scale, and occlusion-aware asymmetry while preserving the efficiency of triplane representations. On SemanticKITTI~\cite{behley2019semantickitti}, GaussianSSC improves Stage~1 occupancy by +1.0\% Recall, +2.0\% Precision, and +1.8\% IoU over state-of-the-art baselines, and improves Stage~2 semantic prediction by +1.8\% IoU and +0.8\% mIoU.
\end{abstract}


\section{Introduction}


Scene understanding plays a critical role in many domains, including small-robot navigation~\cite{alverhed2024autonomous, chen2021adoption}, autonomous driving~\cite{behley2019semantickitti, nahavandi2025comprehensive}, and robot localization~\cite{lowry2015visual, liang2025cscpr, liang2024poco}. However, it remains challenging due to the complexity of sensor modalities~\cite{Li2023VoxFormer, liang2025cscpr, liang2024poco, shan2020lio}, the choice of environment representations~\cite{huang2025gaussianformer, Li2023VoxFormer}, and the incompleteness of observations for capturing a full scene~\cite{dou20203d, Liang2024ETFormer}. Semantic Scene Completion (SSC)~\cite{Cao2022MonoScene} addresses these challenges by generating a dense 3D semantic occupancy map from partial observations (e.g., a single RGB image), filling in occluded and out-of-FOV regions~\cite{Roldo20213DSS,Song2017SSCNet}. Camera-only SSC is particularly attractive because of its lower cost and compactness compared to LiDAR-based SSC, while also providing richer color semantics in addition to geometric cues. Nonetheless, camera-only SSC remains difficult due to the limited field of view~\cite{Xu2024ASO, Liu2022BEVFusionMM,Hou2024FastOccA3} and the complexity of real-world outdoor environments, such as dynamic objects and diverse terrains.

Despite recent progress in monocular semantic scene completion (SSC), which has pushed accuracy while reducing the memory on standard benchmarks~\cite{Liang2024ETFormer, Song2017SSCNet, Li2023VoxFormer}, two issues persist. First, monocular SSC must simultaneously infer semantics and lift them precisely into 3D. From a single view, errors propagate along camera rays and distort geometry, while estimating occluded or out-of-FOV regions is particularly difficult~\cite{Li2023VoxFormer,Huang2023TPVFormer}. Second, grid-centric approaches typically propagate features on uniform lattices, whereas real scenes are anisotropic (e.g., roads, façades, elongated objects). Voxel- or plane-based grids allocate computation uniformly, even to empty space, which creates redundancy and limits adaptivity to scale and structure~\cite{Li2022BEVFormer, Yu2023FlashOcc}.

These limitations motivate the use of Gaussian representations for 3D environments~\cite{Kerbl20233DGS}, since Gaussian distributions are more flexible in modeling the uncertainties of semantic estimation. By attaching semantics to 3D Gaussians with learned means and covariances, Gaussian distributions can adapt to object scale and regional complexity, reducing redundancy while preserving fine structure. Dense occupancy can then be generated via Gaussian-to-voxel splatting~\cite{Huang2024GaussianFormer,huang2025gaussianformer}. In practice, this local and adaptive representation can effectively capture the surfaces of objects and terrains in real scenes. However, naively adopting full Gaussian pipelines can be computationally expensive and difficult to integrate with grid-based SSC decoders.

\begin{figure}[tp]
    \centering
    \includegraphics[width=\columnwidth]{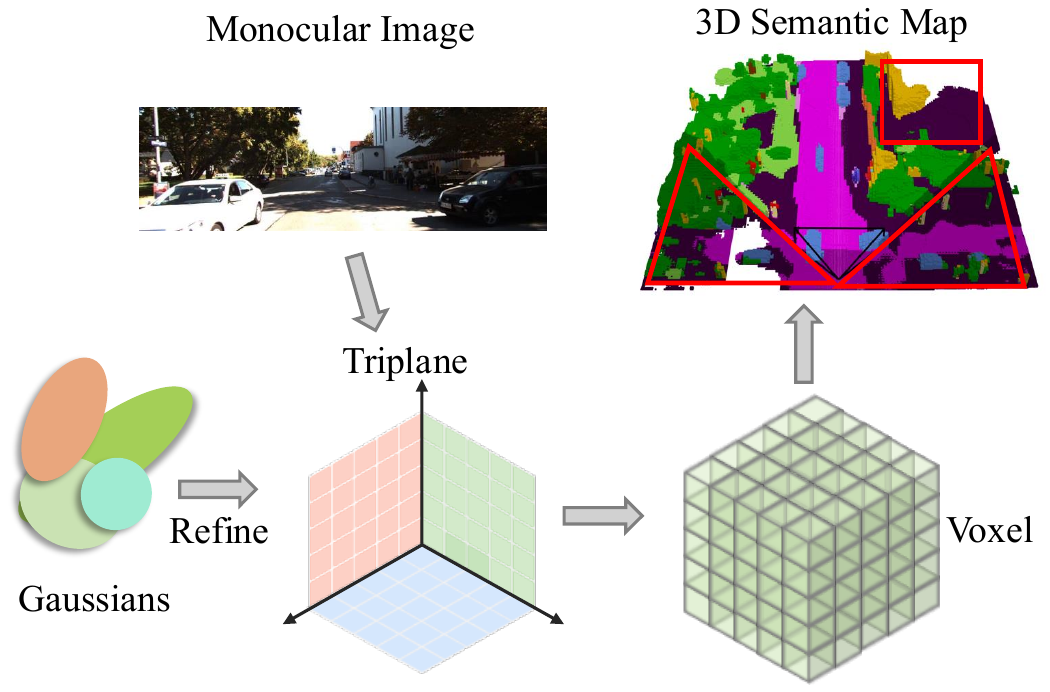}
    \vspace{-2em}
    \caption{Our GaussianSSC takes a monocular input, learns semantics efficiently via a triplane representation, then refines these features with Gaussian primitives to capture geometry and fine details, before fusing into a voxel space for 3D semantic scene completion.}
    \label{fig:cover} 
    \vspace{-2em}
\end{figure}
\textbf{Main Results:} We introduce \emph{GaussianSSC}, a two-stage grid‑native, triplane‑guided SSC framework that integrates Gaussian representations without replacing the voxel grid or maintaining a separate Gaussian set. GaussianSSC converts point-like features into a learned field of Gaussians and use it to refine the triplane features (Figure.~\ref{fig:cover}), to capture surface tangency, scale, and occlusion-aware asymmetry, for better occupancy and semantic map generation. This hybrid triplane–Gaussian design preserves triplane efficiency and uncertainty modeling while importing the local, adaptive support characteristic of Gaussians. The main contributions of our work include:

\begin{itemize}
    \item We introduce a hybrid triplane--Gaussian representation that preserves triplane efficiency while endowing each voxel with Gaussian support to capture fine, surface‑aligned details and occlusion asymmetry in a grid‑native manner.
    \item  We propose Gaussian Anchoring in Stage~1, a sub‑pixel Gaussian‑weighted image aggregation on fused FPN features that tightens voxel–image alignment and improves monocular occupancy estimation.
    
    \item We introduce Gaussian--Triplane refinement in Stage~2, where learned 3D Gaussians refine triplane features via local gathering and global aggregation, yielding continuous, structure‑preserving semantic features.

    \item Our pipeline is fully differentiable and grid–native, and outperform current state-of-the-art methods on SemanticKITTI dataset~\cite{behley2019semantickitti}. Specifically, We achieve 1\% improvement on Recall, 2\% improvement in precision and a 1.8\% improvement in IoU compared to the state-of-the-art approaches for stage 1 occupancy map estimation. We observe a 1.8\% improvement and 0.8\% improvement on IoU and mIoU for stage 2 semantic prediction.

\end{itemize}
\section{Related Work}

\subsection{3D Semantic Scene Completion}
Semantic scene completion (SSC) aims to label every voxel in a 3D volume with occupancy and semantics, including occluded and out-of-FOV regions. Early depth-based work (SSCNet) established the joint occupancy+semantics formulation on dense voxel grids using 3D CNNs \cite{Song2017SSCNet}. With monocular input, previous methods lift image features into 3D and decode voxel-wise labels; MonoScene demonstrated an end-to-end RGB to 3D pipeline, but also underscored the difficulty of inferring geometry and semantics from a single view \cite{Cao2022MonoScene}. Transformer-style pipelines can improve the accuracy--efficiency trade-off by reasoning sparsely in 3D and decoupling occupancy from semantics (e.g., VoxFormer and related designs) \cite{Li2023VoxFormer}. Further efficiency and stability have come from plane/grid factorizations and uncertainty-aware query decoding in triplane-query pipelines (e.g., ET-Former) \cite{Liang2024ETFormer}. To reduce cost, recent designs prune or reorganize voxel queries and stage the reconstruction from coarse to fine; however, aggressive upsampling can lose detail and depth-guided filtering can miss occluded regions. BEV-centric pipelines (e.g., tri-view encodings or dual-path transformers) improve image--volume coupling and long-range context, yet height compression and uniform lattices still limit adaptivity to scale and structure~\cite{Huang2023TPVFormer,Zhang2023OccFormer}. Another idea is to represent scenes with 3D semantic Gaussians, attaching learnable means/covariances and semantics to a compact set, then performing Gaussian-to-voxel splatting to obtain dense occupancy \cite{Huang2024GaussianFormer}. 
Our work bridges grid/plane-based monocular SSC and semantic Gaussian representations. Specifically, we learn a per-voxel Gaussian field and introduce a Gaussian--triplane refinement module to refine semantic features both locally and globally, yielding content-adaptive support without abandoning the simplicity and deployability of grid/plane frameworks.

\subsection{3D Gaussian Splatting}

3D Gaussian splatting (3DGS) represents a scene with anisotropic Gaussians and renders by visibility-aware rasterization, achieving real-time, high-fidelity view synthesis while avoiding waste in empty space \cite{Kerbl20233DGS}. Subsequent work addresses aliasing and scale shifts with anti-aliasing filters (Mip-Splatting) \cite{Yu2024MipSplatting} and extends splatting from static to dynamic scenes via native 4D primitives and deformation fields \cite{Wu20244DGS}. For large environments, CityGaussian and successors propose divide-and-conquer training and multi-level detail for city-scale rendering \cite{Liu2024CityGaussian,Liu2025CityGaussianV2}. 3DGS has also been used for mapping and SLAM (e.g., SplaTAM)~\cite{Keetha2024SplaTAM}. These advances establish Gaussians as an explicit, efficient substrate with strong geometric fidelity and controllable spatial support, complementary to dense grids. 

Beyond view synthesis, semantic or vision extensions bind category or open-vocabulary embeddings to Gaussians for scene understanding, including Semantic Gaussians, CLIP-guided Gaussians, and generalizable semantic splatting~\cite{OpenVocab2024SemanticGaussians,CLIPGS2024,GSemSplat2024}. Multi-sensor variants adapt splats to driving scenarios like LiDAR-GS, tightly-coupled LiDAR–camera splatting; SplatAD for real-time sensor rendering, highlighting 3DGS’s flexibility for autonomous systems \cite{LiDARGS2024,TCLCGS2024,Hess2025SplatAD}. Our approach borrows the adaptive, local aggregation formulation, but remains grid-native for SSC. Instead of keeping a separate Gaussian set, we learn a per-voxel Gaussian field and the projections on the triplanes instead of in 3D space to improve the efficiency.
\section{Method}
In this section, we first formulate the problem in Section.~\ref{sec:problem}. We then introduce GaussianSSC, a two-stage framework, detailing Stage 1 and Stage 2 in Section.~\ref{sec:stage1} and Section.~\ref{sec:stage2}. Finally, we describe our training strategy for GaussianSSC in Section.~\ref{sec:training}.


\begin{figure*}[t]
  \centering
  \includegraphics[width=\textwidth]{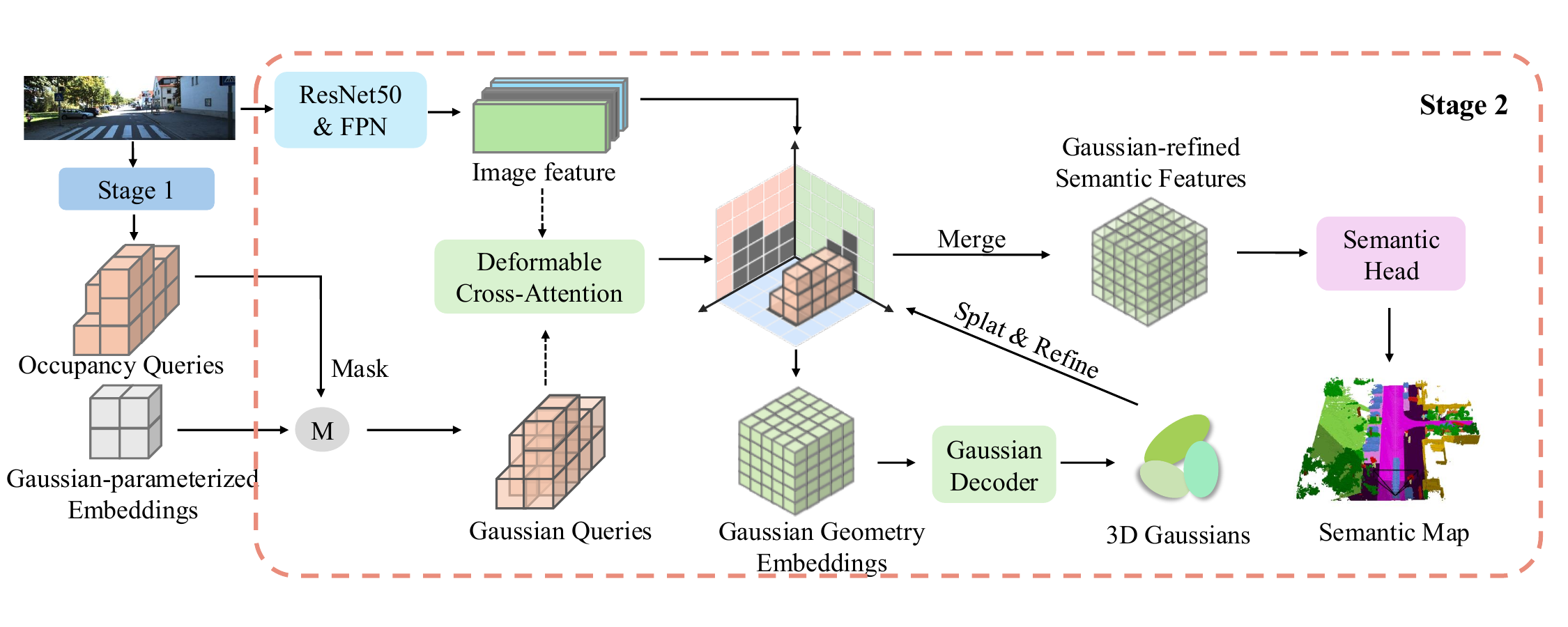}
  \vspace{-3em}
  \caption{\textbf{Overview of Our GaussianSSC:} We present GaussianSSC, a two‑stage pipeline for semantic scene completion. In Stage~1, it predicts an occupancy map from a monocular image, which serves as a structural prior for Stage~2. In Stage~2, we instantiate Gaussian embeddings at voxel locations, gate them by the occupancy priors, and condition three orthogonal triplanes to multi‑scale image features. We then decode a 3D Gaussian per voxel and splat it into the triplanes to perform Gaussian–Triplane refinement to produce stronger semantic features (see Figure~\ref{fig:stage_1_triplane}). The refined triplane features are lifted and merged back to voxel space, where a semantic head predicts the final dense semantic map.}
  \label{fig:pipeline}
  \vspace{-2em}
\end{figure*}
\subsection{Problem Formulation}
\label{sec:problem}
We study monocular semantic scene completion (SSC) from a single RGB image.
Given an input image \(I\in\mathbb{R}^{H\times W\times3}\) with camera intrinsics \(\mathbf{K}\) and extrinsics \(\mathbf{T}=[\mathbf{R}\mid\mathbf{t}]\), the goal is to predict a dense 3D semantic occupancy volume in a predefined region of interest, including occluded and out-of-FOV space beyond the visible surfaces.

\paragraph{Voxelized prediction target.}
We represent the 3D region by a voxel grid \(\mathcal{G}\). For each voxel \(v\in\mathcal{G}\), the model outputs a binary occupancy variable \(o_v\in\{0,1\}\) and a semantic label \(y_v\in\{0,\dots,C{-}1\}\).
The supervision is provided as voxelized ground truth occupancy and semantics.

\paragraph{Image-to-voxel association.}
To relate 2D evidence to 3D voxels, each voxel center \(\mathbf{p}_v\) is mapped to the image plane via the pinhole projection
\[
\mathbf{u}_v=\Pi\!\big(\mathbf{K}[\mathbf{R}\ \mathbf{t}][\mathbf{p}_v^\top,1]^\top\big).
\]
Since the mapping from pixels to depth is underconstrained from a single view and multiple 3D locations may correspond to similar image evidence, monocular SSC is inherently ambiguous, especially in occluded regions.

\paragraph{Two-stage factorization.}
We model SSC with a two-stage factorization that decouples geometry support from semantic labeling:
\[
p_{\theta_1}(o\mid I)\ \text{(Stage 1: occupancy prior)}\]
\[
p_{\theta_2}(y\mid I,o)\ \text{(Stage 2: semantics given structure)}.
\]
Stage~1 estimates an occupancy prior that provides a structural support, while Stage~2 predicts semantics conditioned on this support and the image features.

\subsection{Method Overview}
\label{sec:overview}
GaussianSSC is a two-stage monocular SSC framework that builds semantic completion on top of an explicit occupancy prior.
In Stage~1, we estimate occupancy and extract voxel-aligned features by \emph{Gaussian Anchoring}, which aggregates a learnable local neighborhood on the fused image feature map around each projected voxel location to improve image-to-voxel association.
In Stage~2, we perform semantic completion conditioned on the Stage-1 structure using a triplane-factorized representation, and refine triplane features via \emph{Gaussian--Triplane Refinement} that combines local gathering and global propagation to enhance both boundary fidelity and long-range consistency.
Overall, GaussianSSC leverages triplane factorization for efficient 3D reasoning while introducing Gaussian-conditioned, content-adaptive support to better handle monocular ambiguity and occlusion.

\subsection{Stage 1: Query-Conditioned Triplane with Gaussian Splatting} \label{sec:stage1}
\begin{figure}[tp]
    \centering
    \includegraphics[width=\columnwidth]{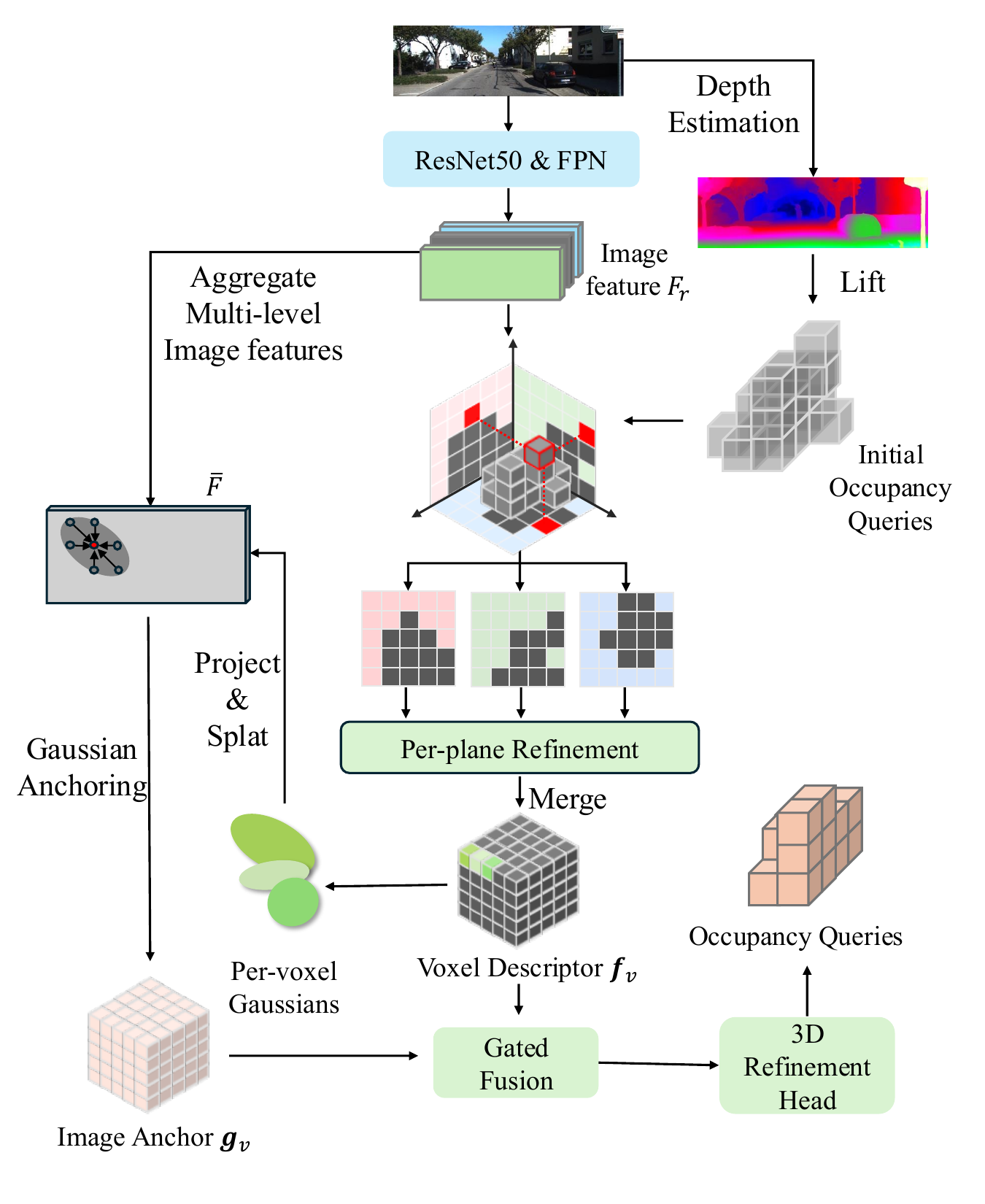}
    \vspace{-3em}
    \caption{\textbf{Illustration of Stage1:
    } Stage 1: From a monocular image, we build a query‑conditioned triplane to obtain voxel descriptors, then apply Gaussian Anchoring—per‑voxel Gaussian windowing on a fused FPN map—to gather sub‑pixel image evidence, fuse it via a gated residual, and predict occupancy with a lightweight 3D head.
    }
    \label{fig:stage_1_arch} 
    \vspace{-2em}
\end{figure}

As shown in Figure.~\ref{fig:stage_1_arch}, stage 1 estimates dense occupancy by (i) constructing voxel descriptors from image features via a query‑conditioned triplane representation and (ii) anchoring each voxel to the image with a learnable Gaussian aggregation on the FPN, followed by gated fusion and light 3D refinement.
\paragraph{Image features and raw voxel queries.}
As shown in Figure.~\ref{fig:stage_1_arch}, given a monocular image \(I\), we extract multi–scale features with a ResNet–50 + FPN, \(\mathcal{F}=\{F_r\}_{r=1}^{L}\). We then form a compact set of voxel queries \(\mathcal{Q}=\{(\mathbf{x}_q,\mathbf{t}_q)\}\) by selecting likely occupied voxel centers in the ROI (e.g., from a coarse monocular depth prior) and initializing each feature \(\mathbf{t}_q\) by sampling the FPN at the projection of \(\mathbf{x}_q\). This produces a geometry–aware seed set without heavy preprocessing.

\paragraph{Triplane construction from queries.}
We factorize the 3D scene using three orthogonal planes
\[
\mathbf{P}^{\text{HW}}\in\mathbb{R}^{H\times W\times d},\quad
\mathbf{P}^{\text{HD}}\in\mathbb{R}^{H\times D\times d},\quad
\mathbf{P}^{\text{WD}}\in\mathbb{R}^{W\times D\times d},
\]
indexed by \(\pi_{\text{HW}}(x,y,z)=(x,y)\), \(\pi_{\text{HD}}(x,y,z)=(x,z)\), \(\pi_{\text{WD}}(x,y,z)=(y,z)\). Learned axis--wise positional embeddings \(\mathbf{e}_x[x],\mathbf{e}_y[y],\mathbf{e}_z[z]\) are fused by a small MLP \(\phi_{\text{PE}}\) to form per--plane codes:
\[
\boldsymbol{\rho}^{\text{HW}}_{x,y}=\phi_{\text{PE}}([\mathbf{e}_x[x];\mathbf{e}_y[y]])\quad (\text{same for HD/WD}).
\]
Each query \((\mathbf{x}_q,\mathbf{t}_q)\) is voxelized to indices \((x_q,y_q,z_q)\) and rasterized onto the planes via learned projections \(\psi\):
\[
\mathbf{P}^{\text{HW}}[x_q,y_q,:] \mathrel{+}= s_{\text{HW}}\,\psi_{\text{HW}}\!\big(\mathbf{t}_q,\boldsymbol{\rho}^{\text{HW}}_{x_q,y_q}\big),
\]
(and analogously for HD/WD), followed by count normalization to mitigate density bias. Lightweight 2D refinement (Conv/FFN and windowed self--attention) produces \(\tilde{\mathbf{P}}^{\text{HW}},\tilde{\mathbf{P}}^{\text{HD}},\tilde{\mathbf{P}}^{\text{WD}}\). Finally, we gather plane features at voxel \((x,y,z)\) and merge (sum/concat+MLP) to obtain voxel--aligned descriptors:
\[
\mathbf{f}_v=\phi_{\text{merge}}\!\big(\tilde{\mathbf{P}}^{\text{HW}}[x,y,:],~\tilde{\mathbf{P}}^{\text{HD}}[x,z,:],~\tilde{\mathbf{P}}^{\text{WD}}[y,z,:]\big)\in\mathbb{R}^{d}.
\]

This triplane factorization provides wide receptive fields at the cost of 2D processing, making 3D reasoning tractable without sacrificing spatial context.

\paragraph{Gaussian Anchoring.}
To robustly associate image evidence with each voxel, we first fuse all FPN levels \(\{F_r\}\) into a single feature map \(\bar{F}\) at a common resolution (e.g., learned per‑level weights). For a voxel center \(\mathbf{p}_v\) with image projection \(\mathbf{u}_v=\Pi(\mathbf{K}[\mathbf{R}~\mathbf{t}][\mathbf{p}_v^\top,1]^\top)\) and fused‑FPN coordinate \(\mathbf{u}'_v\), we treat the voxel as a latent 3D Gaussian whose image footprint is a 2D Gaussian near \(\mathbf{u}'_v\); rather than explicitly predicting a 3D covariance and projecting it via the local Jacobian, we learn the image‑plane Gaussian \((\boldsymbol{\mu}_v,\boldsymbol{\Sigma}_{v})\) directly on \(\bar{F}\), which can be viewed as modeling the projected 3D footprint up to the local projection Jacobian and depth scaling. From the voxel descriptor \(\mathbf{f}_v\) we predict
\[
\boldsymbol{\mu}_v=\mathbf{u}'_v+\boldsymbol{\delta}_v,\qquad
\boldsymbol{\Sigma}_{v}=\mathrm{diag}(\sigma_{x,v}^2,\sigma_{y,v}^2),\qquad
\alpha_v\in(0,1],
\]
and compute a weighted image anchor feature $\mathbf{g}_v$ over a fixed neighborhood \(\mathcal{W}(\boldsymbol{\mu}_v)\):
\[
\tilde{w}_{v,ij}=\alpha_v \exp\!\Big(-\tfrac{1}{2}(\mathbf{u}_{ij}-\boldsymbol{\mu}_v)^\top \boldsymbol{\Sigma}_{v}^{-1}(\mathbf{u}_{ij}-\boldsymbol{\mu}_v)\Big)\]
\[
w_{v,ij}=\frac{\tilde{w}_{v,ij}}{\sum_{(a,b)\in\mathcal{W}(\boldsymbol{\mu}_v)}\tilde{w}_{v,ab}},\quad
\mathbf{g}_v=\sum_{(i,j)\in\mathcal{W}(\boldsymbol{\mu}_v)} w_{v,ij}\,\bar{F}(:,i,j).
\]
 By learning offsets \(\boldsymbol{\delta}_v\), scales \(\boldsymbol{\Sigma}_v\), and opacity \(\alpha_v\), the splat adapts to texture, perspective, and calibration noise, providing sub-pixel alignment and content-aware support. Instead of sampling a single feature at the projected voxel location, Gaussian anchoring aggregates a learnable local neighborhood on the image feature map, which makes image-to-voxel lifting more robust to sub-pixel projection, discretization, and monocular depth ambiguity.
This content-adaptive support reduces spurious activations caused by misalignment and improves the reliability of the occupancy prior.

\paragraph{Gated fusion and 3D refinement.}
Given the voxel descriptor \(\mathbf{f}_v\) and the image anchor \(\mathbf{g}_v\), we apply a gated residual fusion before volumetric refinement:
\[
\mathbf{a}_v=\sigma\!\big(\phi_{\text{gate}}([\mathbf{f}_v;\,\phi_{\text{proj}}(\mathbf{g}_v)])\big),\qquad
\tilde{\mathbf{h}}_v=\mathbf{f}_v+\mathbf{a}_v\odot \phi_{\text{proj}}(\mathbf{g}_v),
\]
where \(\phi_{\text{proj}}\) and \(\phi_{\text{gate}}\) are \(1{\times}1{\times}1\) convolutions, \(\sigma\) is a sigmoid, and \(\odot\) denotes elementwise product. We then apply a lightweight 3D head (dilated residual blocks with optional squeeze–excite) over a local neighborhood \(\mathcal{N}(v)\) to obtain occupancy:
\[
p(o_v{=}{0,1})=\mathrm{softmax}(\text{Head3D}\!\big(\{\tilde{\mathbf{h}}_u\}_{u\in\mathcal{N}(v)}\big)\in\mathbb{R}^{2}).
\]
This gating adaptively injects image evidence where projection is reliable and suppresses it where noisy, while the 3D head enforces local volumetric consistency.

\subsection{Stage 2: Triplane-Conditioned Semantic Completion with Gaussian Splatting} \label{sec:stage2}
Stage~2 estimates dense semantics by (i) conditioning three orthogonal triplanes via self-attention and deformable cross-attention to multi-scale image features, and (ii) applying a triplane-aligned Gaussian refinement that refines voxel features and yields sharper, more coherent semantic maps (Fig.~\ref{fig:pipeline}).

\paragraph{Occupancy-gated Gaussian-parameterized tokens and image conditioning.}
We model each voxel embedding as a \emph{Gaussian-parameterized} token, endowing it with learnable, geometry-aware spatial support (decoded later as directional extents). 
Let $\mathcal{G}$ be the voxel grid and $M_v\!\in\!\{0,1\}$ the Stage-1 occupancy for voxel $v$. 
We instantiate an embedding $e_v\!\in\!\mathbb{R}^D$ at every voxel and gate it by occupancy,
\[
e_v^{\mathrm{occ}} \;=\; M_v \cdot e_v,
\]
and denote the active set $\mathcal{Q}^{\mathrm{G}}_{\mathrm{occ}}=\{\,e_v^{\mathrm{occ}}\;|\;M_v=1\,\}$ as the \emph{Gaussian queries}. 
These queries attend to multi-scale image features $\{F_r\}$ (ResNet-50+FPN) via deformable cross-attention at geometry-aware reference points, producing image-conditioned tokens $e'_v$ that carry salient appearance cues while preserving the Gaussian-parameterized interpretation for downstream aggregation.

\begin{figure}[tp]
    \centering
    \includegraphics[width=0.9\columnwidth]{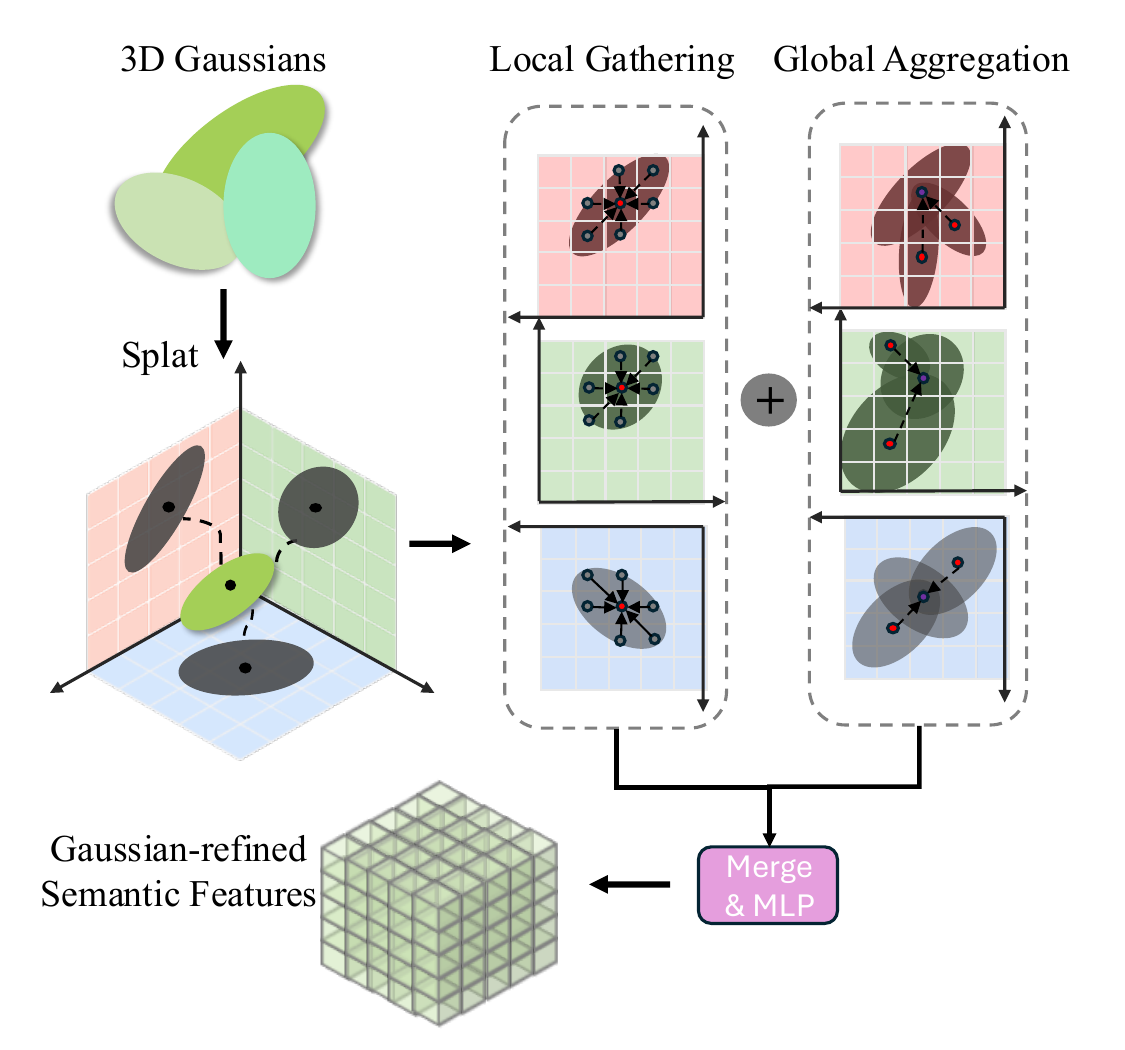}
    \vspace{-1em}
    \caption{\textbf{Illustration of Gaussian-Triplane Refinement:
    } We associate each voxel with a Gaussian centered at the voxel and project this Gaussian onto the three orthogonal triplanes. On each plane, we refine the feature at the projected mean with two complementary steps: (i) local gathering, which anchors the feature by aggregating neighboring evidence within the Gaussian field; and (ii) global aggregation, which shares semantic information from all other locations whose Gaussians cover that point.}
    \label{fig:stage_1_triplane} 
    \vspace{-2em}
\end{figure}

\paragraph{Triplane construction and per-plane processing.}
Following Stage~1, we lift the occupancy-gated, image-conditioned tokens $\{e'_v\}$ onto three orthogonal planes 
$\mathbf{P}^{\mathrm{HW}}, \mathbf{P}^{\mathrm{HD}}, \mathbf{P}^{\mathrm{WD}}$
using learned axis-wise positional embeddings and scatter–add. 
Each plane is refined by lightweight 2D blocks: 
(i) self-deformable attention to propagate information along the plane and extrapolate structure from occupied seeds into nearby unobserved regions, and 
(ii) per-plane deformable cross-attention to $\{F_r\}$ to inject appearance. 
This yields image-conditioned planes 
$\tilde{\mathbf{P}}^{\mathrm{HW}}, \tilde{\mathbf{P}}^{\mathrm{HD}}, \tilde{\mathbf{P}}^{\mathrm{WD}}$.
We then broadcast along the missing axis and merge the three planes at voxel $(x,y,z)$ (sum or concat+MLP) to obtain a voxel-aligned
\emph{Gaussian-geometry embedding} $g_v^{G}$, which concentrates cues predictive of local geometric support (mean-centered, directional extents).

\paragraph{Gaussian–Triplane refinement: local gathering and global aggregation.}
We conceptually assign a 3D Gaussian to each voxel center but avoid full 3D splatting by working per plane. 
From $g_v^{G}$ we decode \emph{three} 2D Gaussian parameter sets—one per triplane $P\!\in\!\{\mathrm{HW},\mathrm{HD},\mathrm{WD}\}$—and a shared opacity $\alpha_v\!\in\!(0,1)$. 
For a plane $P$, let $W_P(\cdot;\theta_v)$ denote the normalized 2D Gaussian window centered at $v$’s plane coordinate with parameters $\theta_v$ (directional extents), and let $\mathcal{N}_R(\cdot)$ be a bounded neighborhood induced by those extents.

\textit{(1) Local gathering (target-centric).} 
Each target location $i$ gathers nearby evidence from $\tilde{\mathbf{P}}$ using a Gaussian window anchored at $i$:
\[
\hat{\mathbf{P}}^{\,\mathrm{gather}}[i]
\;=\; 
\frac{\sum_{j\in\mathcal{N}_R(i)} W_P(i\!-\!j;\theta_i)\,\tilde{\mathbf{P}}[j]}
     {\sum_{j\in\mathcal{N}_R(i)} W_P(i\!-\!j;\theta_i)}.
\]
This improves localization, reduces quantization artifacts, and aligns features to the triplane geometry.

\textit{(2) Global aggregation (source-centric).} 
Each source location $j$ distributes its feature to neighbors according to its own Gaussian and opacity:
\[
\hat{\mathbf{P}}^{\,\mathrm{agg}}[i]
\;=\; 
\frac{\sum_{j\in\mathcal{N}_R(i)} \alpha_j\,W_P(i\!-\!j;\theta_j)\,\tilde{\mathbf{P}}[j]}
     {\sum_{j\in\mathcal{N}_R(i)} \alpha_j\,W_P(i\!-\!j;\theta_j)}.
\]
Thus, when a voxel is covered by multiple Gaussians, their contributions are naturally \emph{normalized and averaged} at that location.

We blend the two refinements per plane with $\beta\!\in\![0,1]$:
\[
\hat{\mathbf{P}} \;=\; \beta\,\hat{\mathbf{P}}^{\,\mathrm{gather}} \;+\; (1{-}\beta)\,\hat{\mathbf{P}}^{\,\mathrm{agg}}.
\]
Finally, we lift/broadcast the three refined planes back to 3D and merge them (sum or concat+MLP) to produce the
\emph{Gaussian-refined voxel feature} $\tilde{\mathbf{h}}_v$, which is fed to the semantic head for the dense semantic map.

\subsection{Training Strategy} \label{sec:training}
\paragraph{Stage~1 (occupancy).}
We train the occupancy head with class–balanced cross–entropy over all valid voxels. Let \(\mathbf{z}^{(1)}_v\in\mathbb{R}^{2}\) be the binary logits and \(p^{(1)}(o_v{=}1)=\mathrm{softmax}(\mathbf{z}^{(1)}_v)_1\). With class weights \(w_0,w_1>0\) (e.g., \(w_0=1-\alpha,~w_1=\alpha\)), the loss is

\[
\begin{aligned}
\mathcal{L}_{\text{CE}}
&= - \sum_{v\in\mathcal{V}} \Big(
w_1\,\mathbb{1}[o_v{=}1]\log p^{(1)}(o_v{=}1) \\
&\qquad\qquad
+\, w_0\,\mathbb{1}[o_v{=}0]\log\big(1 - p^{(1)}(o_v{=}1)\big)
\Big).
\end{aligned}
\]

To stabilize the learned image kernels, we regularize scales and offsets with a weak prior:
\[
\mathcal{L}_{\sigma}=\sum_{v}\big\|\log\boldsymbol{\sigma}_v-\log\boldsymbol{\sigma}_0\big\|_2^2,\qquad
\mathcal{L}_{\delta}=\sum_{v}\|\boldsymbol{\delta}_v\|_1,
\]
where \(\boldsymbol{\sigma}_0\) is a small reference scale. The overall Stage~1 objective is
\[
\mathcal{L}_{\text{stage1}}
=\mathcal{L}_{\text{CE}}
+\lambda_{\sigma}\mathcal{L}_{\sigma}
+\lambda_{\delta}\mathcal{L}_{\delta}.
\]

Furthermore, to mitigate class imbalance without inflating compute, we utilize negative sampling strategy. Specifically, we sample a fixed ratio of negatives per batch and compute the loss over positives plus the sampled negatives. This preserves training stability and focuses gradients on informative empty voxels.

\paragraph{Stage~2 (semantics).}
We optimize class-weighted cross-entropy on valid voxels:
\[
\mathcal{L}_{\text{CE}}=\frac{1}{|\mathcal{V}|}\sum_{v\in\mathcal{V}} w_{y_v}\,\mathrm{CE}\!\big(\mathrm{softmax}(\mathbf{z}^{(2)}_v),y_v\big),
\]
and add light structure-aware penalties that bias volumetric precision, recall, and specificity toward 1:
\[
\begin{aligned}
\mathcal{L}_{\text{sem\_scal}}
&= \frac{1}{C}\sum_{c=1}^{C} \Big(
\operatorname{BCE}(\mathrm{precision}_c,1) + \operatorname{BCE}(\mathrm{recall}_c,1) \\
&\qquad\qquad\qquad
+ \operatorname{BCE}(\mathrm{specificity}_c,1)
\Big).
\end{aligned}
\]

and the composite objective loss
\[
\mathcal{L}_{\text{stage2}}=\lambda_{\text{CE}}\mathcal{L}_{\text{CE}}+\lambda_{\text{sem}}\mathcal{L}_{\text{sem\_scal}}
\]
\section{Results}
We first describe the experimental setup (dataset, metrics, and hardware). We then report quantitative comparisons and qualitative results.

\subsection{Dataset, metrics, and implementation details}
\textbf{Hardware.} We use an NVIDIA RTX A5000 GPU (Intel Xeon(R) W-2255 CPU) for training and evaluation, and report the efficiency benchmark on an RTX A6000 (Table~\ref{tab:efficiency}).

\textbf{Dataset.} We evaluate on SemanticKITTI~\cite{behley2019semantickitti} with the official train/val/test splits. Following common practice, we define a fixed region of size \(51.2\times51.2\times6.4\) m in the LiDAR frame and discretize it into a \(256\times256\times32\) voxel grid at \(0.2\) m resolution. SemanticKITTI comprises 10 sequences for training, 1 sequence for validation, and 11 sequences for testing. It furnishes RGB images with shapes of 1226×370 as inputs and encompasses 20 semantic classes.

\textbf{Implementation details.}
Camera intrinsics/extrinsics are from KITTI calibration files, and images are cropped/resized to match the backbone input.
Sparse voxel queries are derived from dataset-provided geometry together with Stage~1 priors, and rasterized onto triplanes to initialize sparse tokens.
For Gaussians, we set $\boldsymbol{\delta}_v{=}\mathbf{0}$, use $\sigma=\mathrm{softplus}(\hat{\sigma})$ with clamping, and bound $\alpha_v$ by a sigmoid; $\mathcal{W}(\boldsymbol{\mu}_v)$ is a fixed $5{\times}5$ window. For ablations, we use a triplane-query monocular SSC baseline and keep the backbone and training protocol consistent with ETFormer~\cite{Liang2024ETFormer} for fair comparison.

\textbf{Metrics.} We follow the standard SemanticKITTI SSC protocol and exclude unknown/out-of-scope voxels. We report completion (occupancy) IoU with precision/recall, and semantic completion per-class IoU and mIoU, where $\mathrm{IoU}=\frac{\mathrm{TP}}{\mathrm{TP}+\mathrm{FP}+\mathrm{FN}}$, $\mathrm{Precision}=\frac{\mathrm{TP}}{\mathrm{TP}+\mathrm{FP}}$, $\mathrm{Recall}=\frac{\mathrm{TP}}{\mathrm{TP}+\mathrm{FN}}$, and $\mathrm{mIoU}=\frac{1}{C}\sum_{c=0}^{C-1}\mathrm{IoU}_c$.

\begin{figure*}[t]
  \centering
  \includegraphics[width=\textwidth]{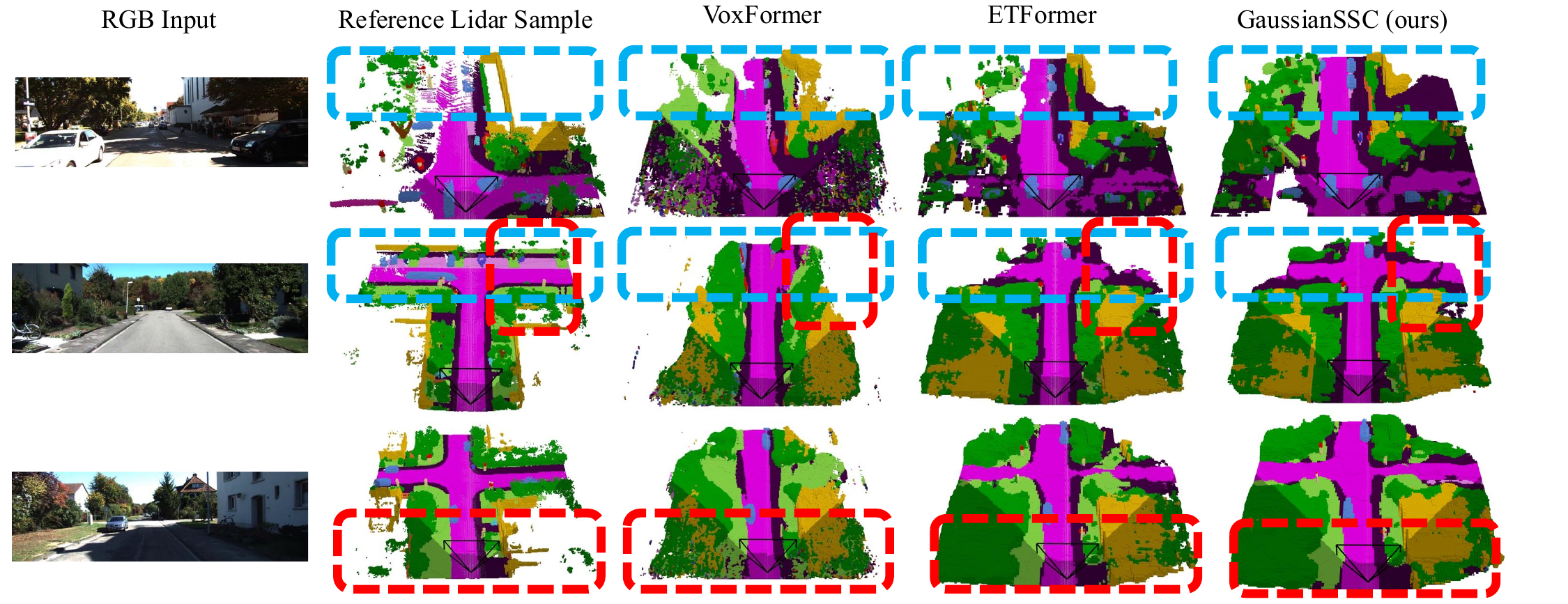}
  \vspace{-2em}
  \caption{\textbf{Visualization results: 
  }
  The figure compares semantic maps generated by different approaches. The reference LiDAR sample is constructed from multiple consecutive LiDAR frames. Blue boxes highlight cases where our method completes the scene more effectively, while red boxes indicate more accurate semantic estimations than other approaches.
  }
  \label{fig:qualitative}
  \vspace{-1em}
\end{figure*}
\subsection{Comparisons}

\begin{table}[!t]
\centering
\resizebox{0.9\columnwidth}{!}{
 \begin{tabular}{c c c c }
                \toprule
        Method & Recall & Precision & IoU \\
                 \midrule
        VoxFormer~\cite{Li2023VoxFormer} & 61.5 & 71.1 & 52.0 \\
        ETFormer~\cite{Liang2024ETFormer} & 71.5 & 77.8 & 59.4 \\
        \textbf{GaussianSSC (Ours)} & \textbf{72.5} (+1.0) & \textbf{79.8} (+2.0) & \textbf{61.2} (+1.8) \\
        \bottomrule 
        \end{tabular}
}
\caption{\textbf{Results of Stage 1 On SemanticKITTI test:
} Our approach, GaussianSSC, outperforms existing methods by at least 1.8\% in IoU and 2\% in precision for occupancy map estimation.}
\vspace{-3em}
\label{tab:stage1}
\end{table}
\paragraph{Stage-1 Occupancy Evaluation}
We evaluate our Stage-1 occupancy predictor (Section~\ref{sec:stage1}) against representative camera-only baselines, including VoxFormer~\cite{Li2023VoxFormer} and ETFormer~\cite{Liang2024ETFormer}.
For fair comparison, we adopt the same depth estimator, MobileStereoNet~\cite{shamsafar2022mobilestereonet}, used in prior monocular SSC pipelines.
As reported in Table~\ref{tab:stage1}, GaussianSSC achieves stronger occupancy estimation performance and yields more reliable occupancy priors.
We attribute the improvement to \emph{Gaussian Anchoring}, which aggregates a learnable local neighborhood around each projected voxel rather than relying on a single pixel sample, making the lifting process more robust to projection quantization and monocular depth ambiguity.
A stronger occupancy prior subsequently reduces the search space for semantic prediction and suppresses false positives in empty regions, leading to more consistent semantic completion in Stage~2.

\begin{table*}[ht]
\scriptsize
    \centering
    \setlength{\tabcolsep}{3pt}
    \begin{tabular}{l|ll|lllllllllllllllllll}
       \toprule
         Methods & IoU & mIoU &  \rotatebox{90}{\fcolorbox{white}{violet!40!magenta}{\makebox[0.1em]{\rule{0pt}{0.1em}}} road } & \rotatebox{90}{\fcolorbox{white}{violet!95!black}{\makebox[0.1em]{\rule{0pt}{0.1em}}} sidewalk} & \rotatebox{90}{\fcolorbox{white}{magenta!40}{\makebox[0.1em]{\rule{0pt}{0.1em}}} parking}&\rotatebox{90}{\fcolorbox{white}{red!40!brown}{\makebox[0.1em]{\rule{0pt}{0.1em}}} other-ground}&\rotatebox{90}{\fcolorbox{white}{yellow!90!black}{\makebox[0.1em]{\rule{0pt}{0.1em}}} building}&\rotatebox{90}{\fcolorbox{white}{blue!50!white}{\makebox[0.1em]{\rule{0pt}{0.1em}}} car}&\rotatebox{90}{\fcolorbox{white}{blue!70!red}{\makebox[0.1em]{\rule{0pt}{0.1em}}} truck}&\rotatebox{90}{\fcolorbox{white}{cyan!30}{\makebox[0.1em]{\rule{0pt}{0.1em}}} bicycle}&\rotatebox{90}{\fcolorbox{white}{blue!80!black}{\makebox[0.1em]{\rule{0pt}{0.1em}}} motorcycle}&\rotatebox{90}{\fcolorbox{white}{blue!70!white}{\makebox[0.1em]{\rule{0pt}{0.1em}}} other-vehicle}&\rotatebox{90}{\fcolorbox{white}{green!75!black}{\makebox[0.1em]{\rule{0pt}{0.1em}}} vegetation}&\rotatebox{90}{\fcolorbox{white}{brown!80!black}{\makebox[0.1em]{\rule{0pt}{0.1em}}} trunk}&\rotatebox{90}{\fcolorbox{white}{green!50!white}{\makebox[0.1em]{\rule{0pt}{0.1em}}} terrain}&\rotatebox{90}{\fcolorbox{white}{red!80!white}{\makebox[0.1em]{\rule{0pt}{0.1em}}} person}&\rotatebox{90}{\fcolorbox{white}{magenta!80}{\makebox[0.1em]{\rule{0pt}{0.1em}}} bicyclist}&\rotatebox{90}{\fcolorbox{white}{brown!60!violet}{\makebox[0.1em]{\rule{0pt}{0.1em}}} motorcyclist}&\rotatebox{90}{\fcolorbox{white}{orange!70!white}{\makebox[0.1em]{\rule{0pt}{0.1em}}} fence}&\rotatebox{90}{\fcolorbox{white}{yellow!30!white}{\makebox[0.1em]{\rule{0pt}{0.1em}}} pole}&\rotatebox{90}{\fcolorbox{white}{red!90!white}{\makebox[0.1em]{\rule{0pt}{0.1em}}} traffic-sign}\\
       \hline
MonoScene~\cite{Cao2022MonoScene} &34.1&11.0&54.7&27.1&24.8&5.7&14.4&18.8&3.3&0.5&0.7&4.4&14.9&2.4&19.5&1.0&1.4&0.4&11.1&3.3&2.1\\ 
TPVFormer~\cite{Huang2023TPVFormer}      &34.2&11.2&55.1&27.2&\underline{27.4}&6.5&14.8&19.2&3.7&1.0&0.5&2.3&13.9&2.6&20.4&1.1&2.4&0.3&11.0&2.9&1.5\\ 
VoxFormer~\cite{Li2023VoxFormer}   &42.9&12.2&53.9&25.3&21.1&5.6&19.8&20.8&3.5&1.0&0.7&3.7&22.4&7.5&21.3&1.4&\underline{2.6}&0.2&11.1&5.1&4.9\\ 
OccFormer~\cite{Zhang2023OccFormer}&34.5&12.3&55.9&\textbf{30.3}&\textbf{31.5}& 6.5&15.7&21.6&1.2&1.5&1.7&3.2&16.8&3.9&21.3&\underline{2.2}&1.1&0.2&11.9&3.8&3.7 \\ 
MonoOcc~\cite{Zheng2024MonoOcc}&-&13.8&55.2&27.8&25.1&\underline{9.7}&21.4&23.2&5.2&\underline{2.2}&1.5&5.4&24.0&8.7&23.0&1.7&2.0&0.2&13.4&5.8&6.4 \\ 
Symphonies~\cite{jiang2024symphonize}&42.1&15.0&\textbf{58.4}&\underline{29.3}&26.9&\textbf{11.7}&24.7&23.6&3.2&\textbf{3.6}&\textbf{2.6}&5.6&24.2&10.0&23.1&\textbf{3.2}&1.9&\textbf{2.0}&\textbf{16.1}&7.7&\textbf{8.0} \\ 
OctreeOcc~\cite{Lu2024OctreeOcc}&44.7&13.1&55.1&26.7&18.6& 0.6&18.6&28.0&\underline{16.4}&0.6&0.7&6.0&25.2&4.8&32.4&\underline{2.2}&2.5&0.0&4.0&3.7&2.3 \\ 
ET-Former~\cite{Liang2024ETFormer}     & \underline{51.4} & \underline{16.3} & 57.6 & 25.8 &16.6 &0.8 &\underline{26.7} &\underline{36.1} & 12.9 &0.6& 0.3 &\underline{8.4} & \underline{33.9} & \underline{11.5} &\underline{37.0} & 1.3&2.5& 0.3&9.5 &\underline{19.6}&6.9 \\ 

\midrule

GaussianSSC (ours) & \textbf{53.2} & \textbf{17.1} & \underline{58.1} & 26.5 & 18.6 & 0.3 & \textbf{28.1}& \textbf{36.4} & \textbf{16.7} &0.75& \underline{1.7} &\textbf{8.5} & \textbf{34.3} & \textbf{12.3} & \textbf{37.7} & 1.3 &\underline{2.8}& 0.0 & 9.5 & \textbf{20.6} & \underline{7.6} \\ 
       \bottomrule 
    \end{tabular}
    \caption{\textbf{Quantitative comparison on SemanticKITTI}~\cite{behley2019semantickitti}.
    GaussianSSC achieves \textbf{53.2/17.1} IoU/mIoU (\textbf{+1.8/+0.8} over prior camera-only best) and yields top accuracy on multiple structural categories (e.g., building, car, truck, vegetation, terrain, and pole).}
    \label{tab:ssc}
    \vspace{-3em}
\end{table*}
\paragraph{Stage-2 Semantics Prediction}
Table~\ref{tab:ssc} reports semantic scene completion results on SemanticKITTI.
GaussianSSC achieves the best overall performance among the compared camera-only methods, indicating improved semantic fidelity and completion quality under monocular ambiguity.
We attribute these gains primarily to the proposed \emph{Gaussian--Triplane Refinement}, which propagates both local and global context on triplanes, together with the Stage-1 occupancy prior that focuses semantic reasoning on plausible occupied regions.
In practice, this combination is particularly beneficial for large planar surfaces and elongated structures, where long-range context and coherent local refinement are crucial.

Figure~\ref{fig:qualitative} provides qualitative comparisons. GaussianSSC produces more complete reconstructions in occluded or out-of-view regions and suppresses spurious predictions in empty space, yielding smoother semantic maps with fewer outliers.
We attribute the reduced noise to the content-adaptive Gaussian support, which aggregates informative local evidence during lifting and refinement.
In some heavily occluded regions, our method may produce denser completions; we note that such behavior is largely aligned with the goal of semantic \emph{scene completion}, i.e., recovering missing structure beyond direct sensor observations.
This reflects a stronger structural prior under monocular ambiguity and entails an inherent trade-off between aggressive completion and conservative empty-space suppression.

\begin{table}[!t]
\centering
\resizebox{0.98\columnwidth}{!}{
 \begin{tabular}{l|cc|cc}
        \toprule
        Method & IoU$\uparrow$ & mIoU$\uparrow$ & Latency (ms)$\downarrow$ & Peak Mem. (GB)$\downarrow$ \\
        \midrule
        VoxFormer~\cite{Li2023VoxFormer} & 50.0 & 14.9 & 309.5 $\pm$ 63.0 & 3.8 \\
        Symphonics~\cite{jiang2024symphonize} & 50.5 & 15.3 & 612.0 $\pm$ 57.8 & 4.3 \\
        ETFormer~\cite{Liang2024ETFormer} & 51.4 & 16.3 & 356.6 $\pm$ 61.7 & 2.3 \\
        \textbf{GaussianSSC (Ours)} & \textbf{53.2} & \textbf{17.1} & 409.8 $\pm$ 56.3 & 2.8 \\
        \bottomrule
 \end{tabular}
}
\caption{\textbf{Accuracy and efficiency on SemanticKITTI.} All results are reported in FP32 with batch size 1 on an RTX A6000.}
\label{tab:efficiency}
\vspace{-3em}
\end{table}
\paragraph{Efficiency Evaluation}
\label{sec:efficiency}
Table~\ref{tab:efficiency} summarizes the accuracy--efficiency trade-off on SemanticKITTI under a unified setting.
VoxFormer attains lower latency but higher peak GPU memory usage due to dense 3D volumetric activations, whereas ETFormer reduces persistent 3D storage via query-based lifting while incurring attention-style runtime overhead.
GaussianSSC introduces moderate additional cost over ETFormer, which is expected from Gaussian-conditioned aggregation and refinement that improve image--voxel alignment under monocular ambiguity.
Symphonics is substantially more expensive in both latency and peak GPU memory usage due to its heavier multi-stage design with scene-encoding instance queries. Overall, GaussianSSC offers a favorable accuracy--efficiency balance among the compared methods.

\paragraph{Ablation study}
\begin{table}[!t]
\centering
\resizebox{0.9\columnwidth}{!}{
 \begin{tabular}{c c c c }
                \toprule
        Method & Recall & Precision & IoU \\
                 \midrule
        Triplane Baseline & 71.5 & 77.8 & 59.4 \\
        \midrule
        + our Gaussian anchoring & 72.2 (+0.7) & 79.4 (+1.6) & 60.8 (+1.4) \\
        + negative sampling & \textbf{72.5} (+0.3) & \textbf{79.8} (+0.4) & \textbf{61.2} (+0.4) \\
        \bottomrule 
        \end{tabular}
}
\caption{\textbf{Stage-1 ablation on SemanticKITTI.}
Compared with the baseline, Gaussian anchoring improves occupancy prediction, and negative sampling provides additional gains.}
\vspace{-3em}
\label{tab:stage1_ablation}
\end{table}
\begin{table}[!t]
\centering
\resizebox{0.9\columnwidth}{!}{
 \begin{tabular}{c c c}
        \toprule
        Method & IoU$\uparrow$ & mIoU$\uparrow$ \\
        \midrule
        Triplane Baseline & 51.4 & 16.3 \\
        \midrule
        + Stage-2 (global-only, $\beta{=}0$) & 52.2 (+0.8) & 16.6 (+0.3) \\
        + Stage-2 (local+global, $\beta{=}0.5$) & 52.6 (+1.2) & 16.8 (+0.5) \\
        + Stage-2 (local-only, $\beta{=}1$) & 52.2 (+0.8) & 16.7 (+0.4) \\
        \midrule
        + Stage-1 \& Stage-2 (local+global, $\beta{=}0.5$) & \textbf{53.2} (+1.8) & \textbf{17.1} (+0.8) \\
        \bottomrule 
 \end{tabular}
}
\caption{\textbf{Stage-2 ablation on SemanticKITTI.} We vary $\beta$ to mix global aggregation ($\beta{=}0$) and local gathering ($\beta{=}1$); $\beta{=}0.5$ performs best. Stage-1 anchoring further boosts completion.}
\vspace{-3em}
\label{tab:stage2_ablation}
\end{table}

We analyze the contribution of each component in GaussianSSC. Table~\ref{tab:stage1_ablation} shows that Gaussian anchoring consistently improves occupancy estimation over the triplane baseline. We attribute this gain to more reliable image-to-voxel alignment: instead of relying on a single projected pixel, the Gaussian window aggregates a learnable local neighborhood, which mitigates projection quantization and monocular ambiguity. Negative sampling further provides additional improvements by exposing the model to hard empty-space negatives, reducing over-confident false positives and yielding a sharper occupancy prior.

Table~\ref{tab:stage2_ablation} demonstrates that the proposed Stage-2 Gaussian--triplane refinement is effective for semantic completion, and that incorporating the Stage-1 occupancy prior further boosts performance. 
Moreover, varying the mixture weight $\beta$ shows that global-only ($\beta{=}0$) and local-only ($\beta{=}1$) refinement each provides partial gains, while combining them ($\beta{=}0.5$) yields the best overall results, indicating that global aggregation and local gathering are complementary. 
This is expected because the occupancy prior focuses semantic reasoning on plausible regions, while the refinement step propagates long-range context and strengthens local consistency, leading to more complete and less noisy semantic occupancy predictions.

\section{Conclusion, Limitations and Future Work}
We presented \emph{GaussianSSC}, a two-stage grid-native framework for monocular semantic scene completion that combines triplane efficiency with Gaussian-conditioned adaptivity.
Gaussian Anchoring improves image-to-voxel alignment for occupancy estimation, while Gaussian--Triplane refinement propagates local--global context for semantic completion within a standard grid/plane representation.
On SemanticKITTI, GaussianSSC achieves consistent gains in both occupancy and semantic completion, demonstrating the complementary benefits of Gaussian support and triplane context.

Despite its effectiveness, Our current formulation operates on single-frame RGB input and does not explicitly model temporal dynamics, which can be important in dynamic scenes.
Moreover, monocular SSC remains inherently ambiguous in severely occluded regions, where multiple completions may be plausible.
We plan to extend the framework to temporal and multi-view settings and to further integrate GaussianSSC into downstream robotics applications such as semantic navigation.

\bibliographystyle{IEEEtran}
\bibliography{root}
\

\end{document}